\documentclass{article}

\usepackage{arxiv}

\usepackage{amsmath,amsfonts}
\usepackage{algorithm}
\usepackage{algpseudocode}

\usepackage[utf8]{inputenc} 
\usepackage[T1]{fontenc}    
\usepackage{hyperref}       
\usepackage{url}            
\usepackage{booktabs}       
\usepackage{amsfonts}       
\usepackage{nicefrac}       
\usepackage{microtype}      
\usepackage{lipsum}
\usepackage{fancyhdr}       
\usepackage{graphicx}       

\usepackage{academicons}
\usepackage{xcolor}

\graphicspath{{media/}}     

\providecommand{\keywords}[1]{\textbf{\textit{Index terms---}} #1}

\newcommand{\orcid}[1]{\href{https://orcid.org/#1}{\textcolor[HTML]{A6CE39}{\aiOrcid}}}

\title{On the Soundness of XAI in Prognostics and Health Management (PHM)
}

\author{
  David Solís-Martín, Juan Galán-Páez \\
  Departamento de Ciencias de la Computación e Inteligencia Artificial  \\
  Datrik Intelligence, S.A. \\
  Universidad de Sevilla \\
  Sevilla\\
  \texttt{\{david.solis, juan.galan\}@datrik.com} \\
  \texttt{\{dsolis, juangalan\}@us.es} \\
   \And
  Joaquín Borrego-Díaz \\
  Universidad de Sevilla \\
  Sevilla\\
  \texttt{\{jborrego\}@us.es} \\
}

\begin{document}
\maketitle

\begin{abstract}
The aim of Predictive Maintenance, within the field of Prognostics and Health Management (PHM), is to identify and anticipate potential issues in the equipment before these become critical. The main challenge to be addressed is to assess the amount of time a piece of equipment will function effectively before it fails, which is known as Remaining Useful Life (RUL). Deep Learning (DL) models, such as Deep Convolutional Neural Networks (DCNN) and Long Short-Term Memory (LSTM) networks, have been widely adopted to address the task, with great success. However, it is well known that this kind of black box models are opaque decision systems, and it may be hard to explain its outputs to stakeholders (experts in the industrial equipment). Due to the large number of parameters that determine the behavior of these complex models, understanding the reasoning behind the predictions is challenging. This work presents a critical and comparative revision on a number of XAI methods applied on time series regression model for PM. The aim is to explore XAI methods within time series regression, which have been less studied than those for time series classification. The model used during the experimentation is a DCNN trained to predict the RUL of an aircraft engine. The methods are reviewed and compared using a set of metrics that quantifies a number of desirable properties that any XAI method should fulfill. The results show that GRAD-CAM is the most robust method, and that the best layer is not the bottom one, as is commonly seen within the context of Image Processing.
\end{abstract}


\section{Introduction}


Incipient AI systems, as small decision trees, were straightly interpretable but had limited capabilities. Nevertheless, during the last years, the notable increase in the performance of predictive models (for both classification and regression) has been accompanied by an increase in model complexity. This has been at the expense of losing the understanding capacity of the reasons behind each particular prediction. This kind of model is known as \emph{black-box} \cite{ref-pomerleau} due to the opaqueness of its behavior. Such obscurity becomes a problem, especially when the predictions of a model impact different dimensions within the human realm (such as Medicine, Law, Profiling, autonomous driving, or Defense, among others) \cite{ref-goodman2017european}. It is also important to note that opaque models are difficult to debug, as opposed to the interpretable ones, which facilitate the detection of the source of its errors/bias and the implementation of a solution \cite{ref-gilpin}.

\subsection{Explainable Artificial Intelligence}

Explainable AI (XAI) addresses these issues by proposing machine learning (ML) techniques that generate explanations of black-box models or create more transparent models (in particular for \emph{post-hoc} explainability). \cite{ref-arieta}. Post-hoc explainability techniques can be divided into model-agnostic and model-specific techniques. Model-agnostic techniques encompass those that can be applied to any ML model, such as LIME \cite{ref-ribeiro} or SHAP \cite{ref-lundberg} (SHapley Additive exPlanations), for example. Whereas model-specific techniques are designed for certain ML models, such as GRAD-CAM (Gradient-weighted Class Activation Mapping) \cite{ref-selvaraju}, Saliency Maps \cite{ref-simonyan}, or Layer-wise Relevance Propagation (LRP) \cite{ref-bach}, which are focused on deep learning (DL) models.

Regarding the kind of tasks where XAI can be applied, it is common to find applications in classification tasks with tabular and image data, while regression tasks -signal processing, among others- have received little attention. The higher number of works devoted to XAI for classification tasks is due to the ease of its application, since implicit knowledge exists around each class \cite{ref-letzgus}. Similarly, there are not many works on XAI applied to time series models \cite{ref-schlegel}. The non-intuitive nature of time series \cite{ref-siddiqui} makes them harder to be understood. Even nowadays, to the best of our knowledge, it seems that there is no specific XAI method particularly successful for this kind of data.

In the same way that it does not exist a model best suited to solve any ML task, there is no particular XAI method that will provide the best explanation of any model. A significant body of literature devoted to innovations in novel interpretable models and explanation strategies can be found. However, quantifying the correctness of their explanations remains challenging. Most ML interpretability research efforts are not aimed at comparing the explanation quality (measuring it) provided by XAI methods \cite{ref-carvalho} \cite{ref-vollert}. It can find two types of \emph{indicators} for the assessment and comparison of explanations: qualitative and quantitative. Quantitative indicators, which are the focus of this paper,  are designed to measure desirable characteristics that any XAI method should have. The metrics approximate the level of accomplishment of each characteristic, thus allowing us to measure them on any XAI method. As these metrics are a form to estimate the accomplishment level, they will be referred to as \emph{proxies}. Numerical proxies are useful to assess the explanation quality, providing a straightforward way to compare different explanations.

A (non-exhaustive) list of works on proxies show its usefulness. In \cite{ref-schlegel}, Schlegel, et al.  apply several XAI methods, usually used with models built from image and text data, and propose a methodology to evaluate them on time series. Samek, at al. \cite{ref-samek} apply a perturbation method on the variables that are important in the prediction generation, to measure the quality of the explanation. The works of Doshi-Velez, et al. and Honegger \cite{ref-doshi} \cite{ref-honegger} propose three proxies (called axioms in those works) for measuring the consistency of explanation methods. The three proxies are identity (identical samples must have identical explanations), separability (non-identical samples cannot have identical explanations) and stability (similar samples must have similar explanations). There exist other works that propose different proxies for more specific models or tasks \cite{ref-silva}.

\subsection{XAI and Predictive Maintenance}

The industrial maintenance process consists of three main stages. The first stage involves identifying and characterizing any faults that have occurred in the system. In the second stage, known as the diagnosis phase, the internal location of the defects is determined, including which parts of the system are affected and what may have caused the faults. In the final stage, known as the prognosis phase, the gathered information is used to predict the machine's operating state, or Remaining Useful Life (RUL) at any given time, based on a description of each system part and its condition. The first two are classification problems, while prognosis is commonly addressed as a regression problem. In the field of predictive maintenance (Pdm), there is an important lack of XAI methods for industrial prognosis problems \cite{ref-vollert} and it is difficult to find existing works in which new XAI methods are developed, or existing ones are applied within this context. Hong, et al. \cite{ref-hong} use SHAP to explain predictions in RUL prediction. Similarly, Szelazek, et al. \cite{ref-szelazek} use an adaptation of SHAP for decision trees applied to steel production systems prognosis, to predict when the thickness of the steel is out of specifications. Serranilla, et al. \cite{ref-serradilla} apply LIME to models used in the estimation of bushings remaining time of life. Recently, Ferrano, et al. \cite{ref-ferrano} have applied SHAP and LIME in hard disk drive failure prognosis.

\subsection{Aim and structure of the paper}

Raw signal time series are frequently voluminous and challenging to analyze data. Due to this issue, a quantitative method must verify the quality of explanations \cite{ref-schlegel}. This paper considers five XAI methods to address regression tasks on signal time series; specifically, for system prognosis within the context of PHM. Two of them, SHAP and LIME, are model-agnostic XAI methods. The other three (layer-wise relevance propagation, gradient activation mapping, and saliency maps) are neural network-specific. It is worth noting that all these methods have been adapted in this paper to work with time series regression models.

This article aims to present several contributions to the field of Explainable Artificial Intelligence. Firstly, it presents a comprehensive review of eight existing proxy methods for evaluating the interpretability of machine learning models. Secondly, it proposes a novel proxy to measure the time-dependence of XAI methods, which has not been previously explored in the literature. Thirdly, an alternative version of Grad-CAM is proposed, which takes into account both the time and time series dimensions of the input, improving its interpretability. Finally, the importance of layers in Grad-CAM for explainability is evaluated using the proposed proxies.

The article is organized into five main sections. Section 2 covers the Materials and Methods used in the study, including the XAI methods and the perturbation and neighborhood techniques. Section 2.3 focuses on the validation of the XAI method explanations using quantitative proxies. Section 3 provides details on the experiments conducted in the study, including the dataset and the black-box model used. Section 3.2 describes the experiments themselves, and Section 3.3 presents the results of the study. Finally, Section 4 provides a discussion of the findings and their implications.

\section{Materials and Methods}

This section describes the different XAI methods under study (Subsect 2.1), as well as the consistency of explanation proxies, used in the experiments carried out (Subsect 2.2).



\subsection{XAI methods} 

This section provides a brief description of the XAI models used in the experiments.

\subsubsection{Local interpretable model-agnostic explanations}

{\em Local Interpretable Model-agnostic Explanations} (LIME) \cite{ref-ribeiro} is a XAI method based on a surrogate model. In XAI, surrogate models are trained to approximate the predictions of the black-box model. These models would be white-box models, easily interpreted (sparse linear models or simple decision trees). In the case of LIME, the surrogate model is trained to approximate an individual prediction and the predictions of its neighborhood obtained by perturbing the individual sample studied. The LIME surrogate model is trained with a data representation of the original sample $x \: \epsilon \: R^d$. The representation uses $x' \: \epsilon  \{ 0,1 \}^{d'}$ to state the non-perturbation/perturbation of each original feature. Mathematically, the explanations obtained with LIME can be expressed as:

\begin{equation}
\xi (x) = \underset{g \: \epsilon \: G}{argmin} \: \mathcal{L}(f, g, \pi_x) + \Omega(g)  
\end{equation}

\noindent where $g$ is a surrogate model from the class $G$ of the all interpretable models. The component $\Omega(g)$ is used as regularization to keep the complexity of $g$ low, since high complexity is opposed to the interpretability concept. The model being explained is denoted as $f$ and $\mathcal{L}$ determines the performance of $g$ fitting the locality defined by $\pi$ as a proximity measurement function and $\pi_x = \pi(x, \cdot )$. Finally, each training sample is weighted with the distance between the perturbed sample and the original sample.

\subsubsection{SHapley Additive exPlanations}

{\em SHapley Additive exPlanations} (SHAP) \cite{ref-lundberg}, is also a method to explain individual predictions, as LIME. The SHAP method explains each feature by computing shapely values from coalitional game theory. Shapley value can be described as the expected average of a player's marginal contribution (by considering all possible combinations). It enables the determination of a payoff for all players, even when each one has contributed differently. In SHAP, each feature is considered a player. Thus, the coalition vector $x'$, or simplified features vector, is composed of 1/0's representing the presence or absence of a feature, respectively. The contribution of each feature $\phi_i$ is estimated based on its \emph{marginal contribution} \cite{ref-shapley}, and computed as follows:

\begin{equation}
    \phi_i(f,x) = \underset{z' \: \subseteq \: x'}{\sum} \frac{\left | z' \right |!(M - \left | z' \right | -1)!}{M!}\left [f_x(z') - f_x(z' \setminus  i \right)]
\end{equation}

\noindent where $\left | z' \right |$ is the number of non-zero entries in $z'$, and $z'  \subseteq  x'$ represents all vectors where the non-zero entries are a subset of the coalition vector $x'$. The values $\phi_i$ are known as Shapely values, and it has been demonstrated that they satisfy the properties of \emph{local accuracy, missingness, and consistency} \cite{ref-lundberg}. Based on these contribution values, a linear model is defined to obtain the explainable model $g$:
\begin{equation}
    g(x') = \phi_o + \underset{j=1}{\overset{M}{\sum}}\phi_j
\end{equation}

The explainable model $g$ is optimized by minimizing the mean squared error between $g$ and the predictions over the perturbed samples $f(h_x(z'))$. The main difference with LIME is that in SHAP, each sample is weighted based on the number of 1's in $z'$. The weighting function, called the weighting kernel, gives rise to the so-called Kernel SHAP method. The formula for the weight $\pi_x(z')$ is given by:

\begin{equation}
\pi_x(z') = \frac{(M-1)}{\binom{M}{\left | z' \right |}(M - \left | z' \right |)}
\end{equation}

The intuition behind this is that isolating features provides more information about their contribution to the prediction. This approach computes only the more informative coalitions, as computing all possible combinations is an intractable problem in most cases.

\subsubsection{Layer-Wise Propagation}

The {\em Layer-wise Propagation} (LRP) method \cite{ref-bach} aims to interpret the predictions of deep neural networks. It is thus a model-specific XAI method. The goal is to attribute relevance scores to each input feature (or neuron) of a neural network, indicating its contribution to the final prediction. LRP works by propagating relevance scores from the output layer of the network back to its input layer. The relevance scores are initialized at the output layer: a score of 1 is assigned to the neuron corresponding to the predicted class and 0 to all others. Then, relevance is propagated backward from layer to layer using a propagation rule that distributes the relevance scores among the inputs of each neuron in proportion to their contribution to the neuron's output. The rule ensures that the sum of relevance scores at each layer is conserved. The propagation rule is defined by the equation.

\begin{equation}
    R_i = \underset{k}{\sum} \frac{z_{ik}}{ \sum_{0,j} z_{jk}} R_k
\end{equation}
where $R$ represents the propagation relevance score, $j$ and $k$ refer to neurons in two consecutive layers, and $z_{jk} = a_j w_{jk}$ denotes how neuron $j$ influences the relevance of neuron $k$ based on the activation of each neuron. The denominator enforces the conservation property.

\subsubsection{Image-Specific Class Saliency}

{\em Image-Specific Class Saliency} \cite{ref-simonyan} is one of the earliest pixel attribution methods existing in the literature. Pixel attribution methods aim to explain the contribution of each individual pixel, within an image, to the model's output. These methods are typically used in computer vision tasks such as image classification or object detection. However, in this work, attribution is assigned to each element of each time series, rather than individual pixels in an image. It is based on approximating the scoring or loss function, $S_c(x)$, with a linear relationship in the neighborhood of x:

\begin{equation}
    S_c(x) \approx w_c^T x + b
\end{equation}
where each element of $w_c$ is the importance of the corresponding element in $x$. The $w_c$ vector of importance values is computed via the derivative of $S_c$ with respect to the input $x$:

\begin{equation}
    w = \frac{\partial S_c}{\partial x} \left. \right|_{x_0}
\end{equation}

This method was originally designed to work in Image Processing with neural networks, hence each element of $w$ is associated with the importance of each pixel.

\subsubsection{Gradient-weighted Class Activation Mapping}

Finally, {\em Gradient-weighted Class Activation Mapping} (Grad-CAM)  generalizes CAM \cite{ref-zhou}, which determines the significance of each neuron in the prediction by considering the gradient information that flows into the last convolutional layer of the CNN. Grad-CAM computes the gradient $y^c$ of class $c$ with respect to a feature map $A^k$ of a convolutional layer, which is then globally averaged, obtaining the neuron importance $\alpha_k^c$ of the feature map $A^k$:

\begin{equation}
    \alpha_k^c = \overset{global \: average \: pooling}{\overbrace{\frac{1}{Z} \underset{i}{\sum}  \underset{j}{\sum}}} \underset{gradients \: via \: backprop}{\underbrace{\frac{\partial y^c}{ \partial A^k}}}
\end{equation}

After computing the importance of all feature maps, a heat map can be obtained through a weighted combination of them. The authors apply a ReLU activation since they are only interested in positive importance;

\begin{equation}
    L^c = ReLU \left ( \underset{k}{\sum}{\alpha_k^c A^k}  \right )
\end{equation}

Unlike saliency maps, GRAD-CAM associates importance by regions of the input. The size of them depends on the size of the convolutional layer. Usually, interpolation is applied to the original heat map to expand it to the overall size of the input.

As this work is focused on time series, we propose introducing additional elements to Grad-CAM with the aim of exploiting the possible stationary information of the signal. This is achieved by introducing an additional component to the Grad-CAM heat map calculation, namely the \textit{time component contribution}. Moreover, a second component was introduced to exploit the importance each time series has in a multivariate time series problem. Thus, the final Grad-CAM attribution equation read as follows, namely the \textit{individual time series contribution}:

\begin{equation}
    \alpha_k^c = \overset{individual \: feature \: contribution}{\overbrace{\frac{1}{T*F} \overset{T}{\underset{i}{\sum}}  \overset{F}{\underset{j}{\sum}} \frac{\partial y^c}{ \partial A^k}}} +  \underset{time \: component \: contribution}{\beta\underbrace{\frac{1}{T} \overset{T}{\underset{i}{\sum}} \frac{\partial y^c}{ \partial A^k}}} + \overset{individual \: time \: series \: contribution}{\overbrace{\sigma \frac{1}{F} \overset{F}{\underset{j}{\sum}} \frac{\partial y^c}{ \partial A^k}}}
\end{equation}
\noindent where $T$ and $F$ are the time units present in the time series, and the number of time series, respectively. The components $\beta$ and $\sigma$ are used to weigh these two new components. 

\begin{figure}
\centering
\includegraphics[width=.95\linewidth]{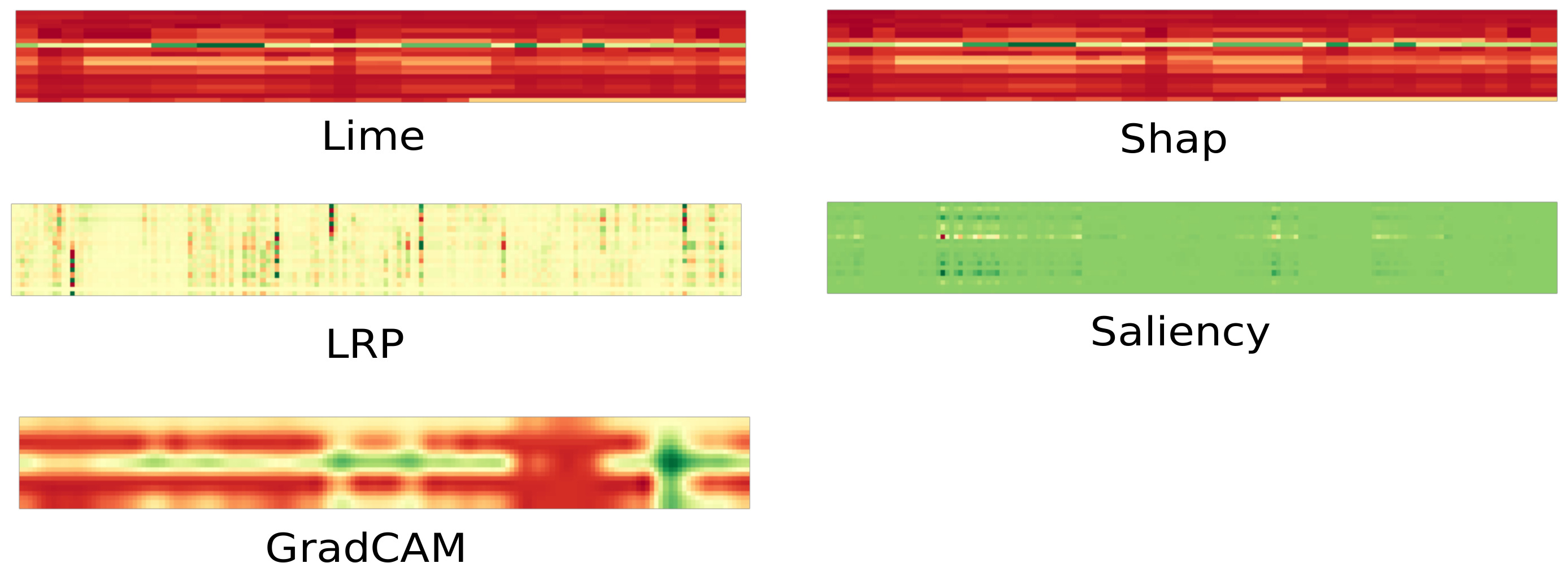}
\caption{Heat maps generated for each of the five tested methods}
\label{fig:heatmaps}
\end{figure}

Figure \ref{fig:heatmaps} displays examples of heat maps generated by each method. Each heat map is a matrix of 20 by 160 values, representing 20 time series and 160 time units, where a relative importance is assigned to each item in the time series.

\subsection{Perturbation and neighborhood}

The LRP, saliency map, and GRAD-CAM techniques can be directly used on time series data. However, LIME and SHAP assume that significant changes in the performance of a \textit{well-trained} model will occur if its relevant features (time points) are altered. Due to the high dimensionality of time series inputs, in order to achieve the former, it is necessary to specify a grouping of time series elements to analyze the impact on each group instead of on single time points. In image processing, this is achieved through super-pixels, which are groups of connected pixels that share a common characteristic. 

\begin{figure}
\centering
\includegraphics[width=.97\linewidth]{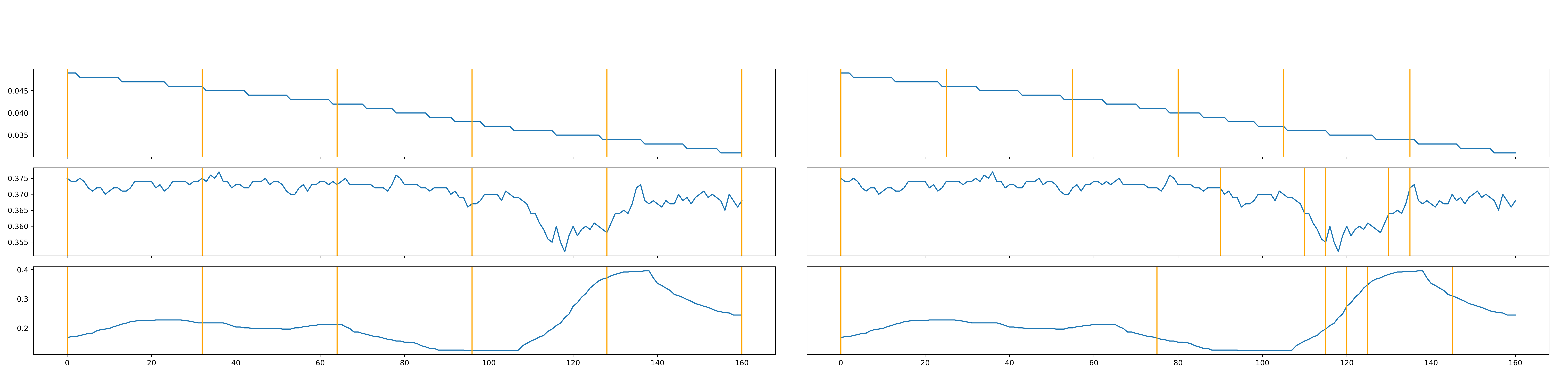}
\caption[Segmentation approaches: Left, uniform segmentation. Right, minimal error segmentation]{Segmentation approaches: Left, uniform segmentation. Right, minimal error segmentation}
\label{fig:splitting}
\end{figure}

In this work, time series are segmented considering adjacent elements. Two different segmentation approaches are used. The first one is called uniform segmentation, which is the most basic method, and involves splitting the time series $ts = {t_0, t_1, t_2, ..., t_n}$ into equally sized windows without overlapping. The total number of windows is $d = \frac{n}{m}$, where $m$ is the size of the window. If $n$ is not divisible by $m$, the last window may be adjusted. The second segmentation minimizes the sum of $l2$ errors by grouping time points as follows:

\begin{equation}
    \epsilon_{l2} (ts_{i,j}) = \overset{j}{\underset{k = i}{\sum}} \left | ts_k - \overline{ts}_{i,j} \right |^2 
\end{equation}
where $ts_{i,j}$ represents a segment of the time series signal $ts$ that goes from element $i$ to element $j$, and $\overline{ts_{i,j}}$ is the mean of that segment. That is to say, the final cost of the segmentation is the sum of the $l2$ errors, calculated between each pair of adjacent elements within the segment. To find the optimal segmentation, a dynamic programming approach is employed, similarly to that described in \cite{ref-truong}. The two segmentation strategies are shown in figure \ref{fig:splitting}.

Once the segmentation is complete, a perturbation method must be applied to create the neighborhood of the time series for SHAP and LIME. The following five perturbation techniques have been applied on the segmented time series, in the different experiments carried out:

\begin{itemize}
    \item Zero: The values in $ts_{i,j}$ are set to zero.
    \item One: The values in $ts_{i,j}$ are set to one.
    \item Mean: The values in $ts_{i,j}$ are replaced with the mean of that segment ($\overline{ts}_{i,j}$).
    \item Uniform Noise: The values in $ts_{i,j}$ are replaced with random noise following a uniform distribution between the minimum and maximum values of the feature.
    \item Normal Noise: The values in $ts_{i,j}$ are replaced with random noise following a normal distribution with mean and standard deviation of the feature.
\end{itemize}

To obtain a perturbed sample $x'$, firstly, it is divided into $n$ segments. Then, from this segmentation, a binary representation $z'$, identifying which segments will be perturbed, is randomly generated:

\begin{equation}
    x' = h(x, z') = \left( h(x,z')_1, h(x,z')_2, ..., h(x,z')_n \right)
\end{equation}
where
\begin{equation}
    h(x, z')_i = \left\{\begin{matrix}
g(x, i) &  if \: z'_i \: is \: equal \: 0\\
p(g(x, i)) & if \: z'_i \: is \: equal \: 1 
\end{matrix}\right. \qquad \qquad i \in \{1,...,n\}
\end{equation}
with $p$ being a perturbation function and $g$ a segmentation function.



\subsection{Validation of XAI method explanations} \label{sect:proxies}

\begin{figure}

\centering
\includegraphics[width=1\linewidth]{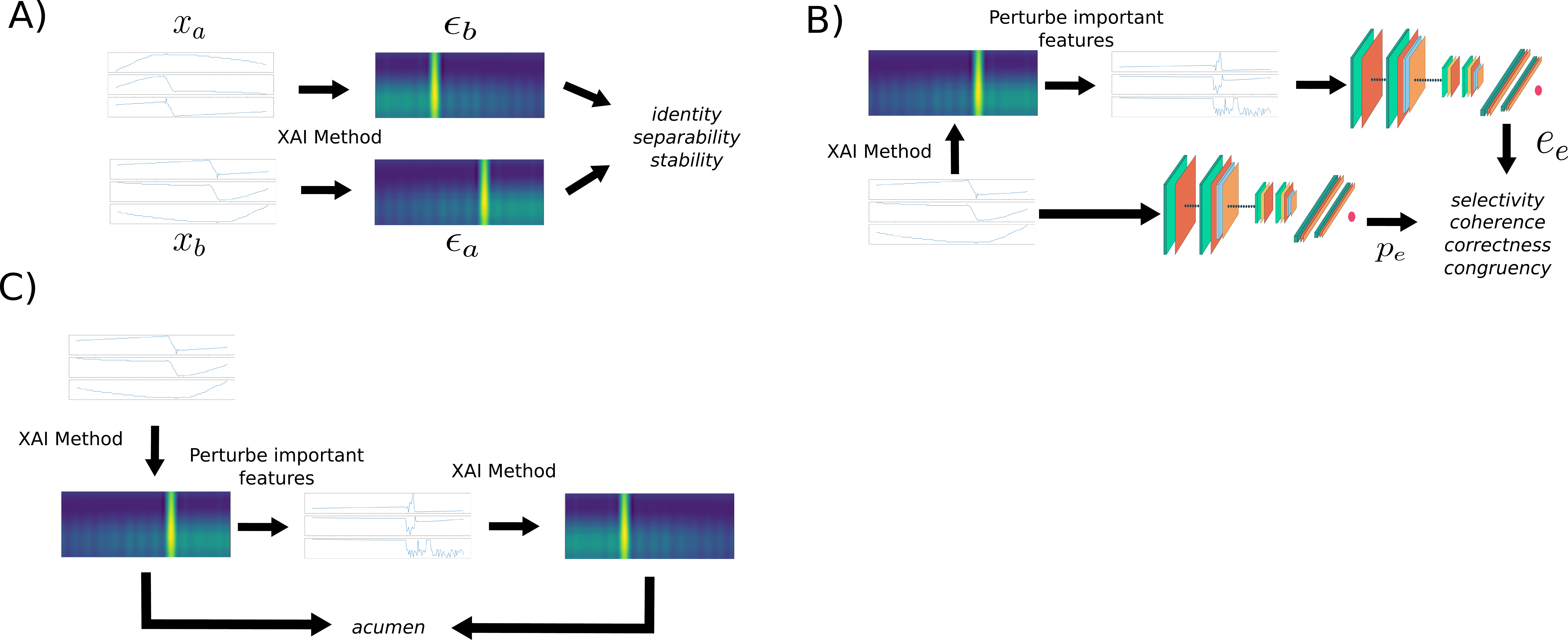}
\caption{Methodologies can be used to compute proxies for evaluating the interpretability of machine learning models.}
\label{fig:cs-method}

\end{figure}

This study uses the most relevant quantitative proxies found in the literature \cite{ref-honegger,ref-rokade}, to evaluate and compare each method. Different methodologies need to be followed depending on the proxy used to evaluate the interpretability of machine learning models. These methodologies are depicted graphically in Figure \ref{fig:cs-method}. Approach A involves using different samples and their corresponding explanations to compute metrics such as identity, separability, and stability. In approach B, a specific sample is perturbed using its own explanation, and the difference in prediction errors (between the original and perturbed sample) is computed. This methodology is used to evaluate metrics such as selectivity, coherence, correctness, and congruency. Approach C involves using the explanation of the original sample to build the perturbed sample, then the explanations from both the original and perturbed samples are used to compute the acumen proxy.

The following are the desirable characteristics that each XAI method should accomplish, and the proxy used for each one in this work: 

\begin{itemize}
    \item \emph{Identity}: The principle of identity states that identical objects should receive identical explanations. This estimates level of intrinsic non-determinism in the method.

    \begin{equation}
        \forall a, b \: \left ( d(x_a, x_b) = 0 \implies d(\epsilon_a, \epsilon_b ) = 0 \right )
    \end{equation}
    $x$ are samples, $d$ is a distance function and $\epsilon$ explanation vectors (which explain the prediction of each sample).
    
    \item \emph{Separability}: Non-identical objects cannot have identical explanations. 
    
  \begin{equation}
         \forall a, b \: \left ( d(x_a, x_b) \neq 0 \implies d(\epsilon_a, \epsilon_b ) > 0 \right )
    \end{equation}   

     If a feature is not actually needed for the prediction, then two samples that differ only in that feature will have the same prediction. In this scenario, the explanation method could provide the same explanation, even though the samples are different.
     For the sake of simplicity, this proxy is based on the assumption that every feature has a minimum level of importance, positive or negative, in the predictions.
    
    \item \emph{Stability}: Similar objects must have similar explanations. This is built on the idea that an explanation method should only return similar explanations for slightly different objects. The Spearman correlation $\rho$ is used to define this:

    \begin{equation}
        \rho ( \left \{   d(x_i, x_0), d(x_i, x_1), ..., d(x_i, x_n) \right \}, \left \{ d(\epsilon_i, \epsilon_0), d(\epsilon_i, \epsilon_1), ..., d(\epsilon_i, \epsilon_n) \right \})  \underset{\forall i}{=} \rho_i > 0 
    \end{equation}  
    
    \item \emph{Selectivity}. The elimination of relevant variables must affect negatively to the prediction \cite{ref-bach} \cite{ref-samek2}. To compute the selectivity, the features are ordered from the most to least relevant. One by one the features are removed, by setting it to zero for example, and the residual errors are obtained to get the area under the curve (AUC). 
    
    \item \emph{Coherence}. It computes the difference between the prediction error $p_e^i$ over the original signal and the prediction error $e_e^i$ of a new signal where the non-important features are removed. 
    
    \begin{equation}
        \alpha_i = \left | p_e^i - e_e^i \right |
    \end{equation}
    where $\alpha_i$ is the coherence of a sample. 

    \item \emph{Completeness}. It evaluates the percentage of the explanation error from its respective prediction error.

    \begin{equation}
        \gamma_i = \frac{e_e^i}{p_e^i}
    \end{equation}

    \item \emph{Congruency}. The standard deviation of the coherence provides the congruency proxy. This metric help to capture the variability of the coherence.
    
    \begin{equation}
        \delta = \sqrt{ \frac{ \sum ((\alpha_i - \overline{\alpha})^2}{N} }
    \end{equation}
    where $\overline{\alpha}$ the average coherence over a set of $N$ samples:
    
    \begin{equation}
        \overline{\alpha} = \frac{\sum \alpha_i}{N}
    \end{equation}
    
    \item \emph{Acumen}. It is a new proxy proposed by the authors for the first time in this paper, based on the idea that an important feature according to the XAI method should be one of the least important after it is perturbed. This proxy aims to detect whether the XAI method depends on the position of the feature, in our case, the time dimension. It is computed by comparing the ranking position of each important feature after perturbing it. 
    
    \begin{equation}
        \varpi = 1 - \frac{ \underset{f_i \in \mathcal{I}}{\sum} \frac{p_a(f_i)}{N}}{M}
    \end{equation}
    where $\mathcal{I}$ is the set of $M$ important features before the perturbation, $p_a(f_i)$ is a function that returns the position of feature $f_i$ within the importances vector after the perturbation, where features with lower importance are located at the beginning of the vector.

\end{itemize}

Some of the previously depicted methods for evaluating the interpretability of machine learning models perturb the most important features identified by the XAI method. In our work, we define the most important features as those whose importance values are greater than 1.5 times the standard deviation of the importance values, up to a maximum of 100 features.



\section{Experiments and results}
\subsection{Dataset and black-box model}

\begin{figure}
\centering
\includegraphics[width=1\linewidth]{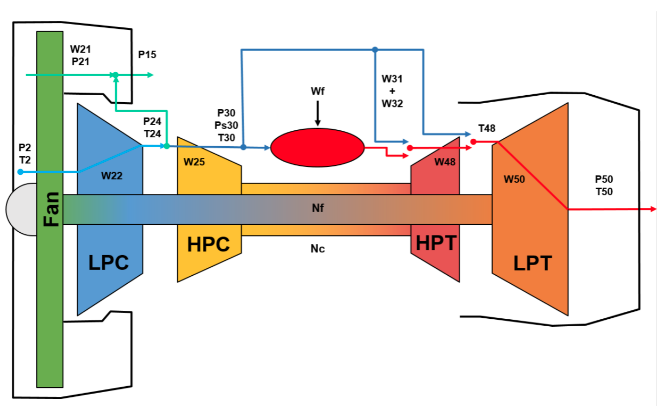}
\caption[Schematic of the model used in N-CMAPSS.]{Schematic of the model used in N-CMAPSS \cite{ref-arias}.}
\label{fig:cmapss_model}
\end{figure}

The Commercial Modular Aero-Propulsion System Simulation (CMAPSS) is a modeling software developed at NASA. It was used to build the well known CMAPSS dataset \cite{ref-saxena} as well as the recently created N-CMAPSS dataset \cite{ref-arias}. N-CMAPSS was created providing the full history of the trajectories starting with a healthy condition until the failure occurs. A schematic of the turbofan model used in the simulations is shown in Figure \ref{fig:cmapss_model}. All rotation components of the engine (fan, LPC, HPC, LPT, and HPT) can be affected by the degradation process. 

Seven different failure modes, related to flow degradation or subcomponent efficiency, that can be present in each flight have been defined. The flights are divided into three classes depending on the length of the flight. Flights with a duration from 1 to 3 hours belong to class 1, class 2 consists of flights between 3 and 5 hours, and flights that take more than 5 hours fall into class 3. Each flight is divided into cycles, covering climb, cruise, and descend operations.

The problem revolves around the development of a model $h$ capable of predicting the remaining useful life $y$ of the system, using the sensor outputs $x_s$, the scenario descriptors $w$ and auxiliary data $a$. The different variables available to estimate the RUL of the system are described in table \ref{table:variables}. The former is an optimization problem that can be denoted as:

\begin{equation}
\underset{h \: \in \: \mathcal{H}}{argmin} \: \sum \mathcal{S} \left ( y -  h(x_s , w, a) \right )
\end{equation}

where $y$ and $h(x_s, w, a)$ are, respectively, the expected and estimated RUL. $\mathcal{H}$ is the set of the different models to be tested by the optimization process and $\mathcal{S}$ is a scoring function defined as the average of the Root-Mean-Square Error (RMSE) and the NASA's scoring function ($N_s$) \cite{ref-saxena}:

\begin{equation}
\mathcal{S} = 0.5 \cdot RMSE + O.5 \cdot N_s
\label{eq:score}
\end{equation}

\begin{equation}
N_s = \frac{1}{M} \sum exp(\alpha | y - \hat{y} |) - 1
\end{equation}

\noindent being $M$ the number of samples and being $\alpha$ equal to $\frac{1}{13}$ in case that $\hat{Y} < Y$ and $\frac{1}{10}$ otherwise.

\begin{table}[]
\centering
\caption{Variable description, symbol, units and variable set.}
\begin{tabular}{llll}
\hline
Symbol & Set & Description                     & Units \\ \hline
alt    & $W$   & Altitude                        & ft    \\
Mach   & $W$   & Flight Mach number              & -     \\
TRA    & $W$   & Throttle-resolver angle         & \%     \\
T2     & $W$   & Total temperature at fan inlet  & ºR    \\
Wf     & $X_s$ & Fuel flow                       & pps   \\
Nf     & $X_s$ & Physical fan speed              & rpm   \\
Nc     & $X_s$ & Physical core speed             & rpm   \\
T24    & $X_s$ & Total temperature at LPC outlet & ºR    \\
T30    & $X_s$ & Total temperature at HPC outlet & ºR    \\
T48    & $X_s$ & Total temperature at HPT outlet & ºR    \\
T50    & $X_s$ & Total temperature at LPT outlet & ºR    \\
P15    & $X_s$ & Total pressure in bypass-duct   & psia  \\
P2     & $X_s$ & Total pressure at fan inlet     & psia  \\
P21    & $X_s$ & Total pressure at fan outlet    & psia  \\
P24    & $X_s$ & Total pressure at LPC outlet    & psia  \\
Ps30   & $X_s$ & Static pressure at HPC outlet   & psia  \\
P40    & $X_s$ & Total pressure at burner outlet & psia  \\
P50    & $X_s$ & Total pressure at LPT outlet    & psia  \\
Fc     & $A$   & Flight class                    & -     \\
$h_s$    & $A$   & Health state                    & -    
\end{tabular}
\label{table:variables}
\end{table}

The model used in the experimentation was designed and implemented by the authors and it achieved the third place in the 2021 PHM Conference Data Challenge \cite{ref-solis}. 

The former 20 variables have different scales, thus a z-score normalization is applied to homogenize the variables scale:

\begin{equation}
    x'_f = \frac{x_f - \mu_f}{\sigma_f}
\end{equation}

\noindent where $x_f$ is the data of a feature $f$, and $\mu_f$ and $\sigma_f$ are its mean and standard deviation, respectively. 

\begin{figure}[t]
\centering
\includegraphics[scale=.3]{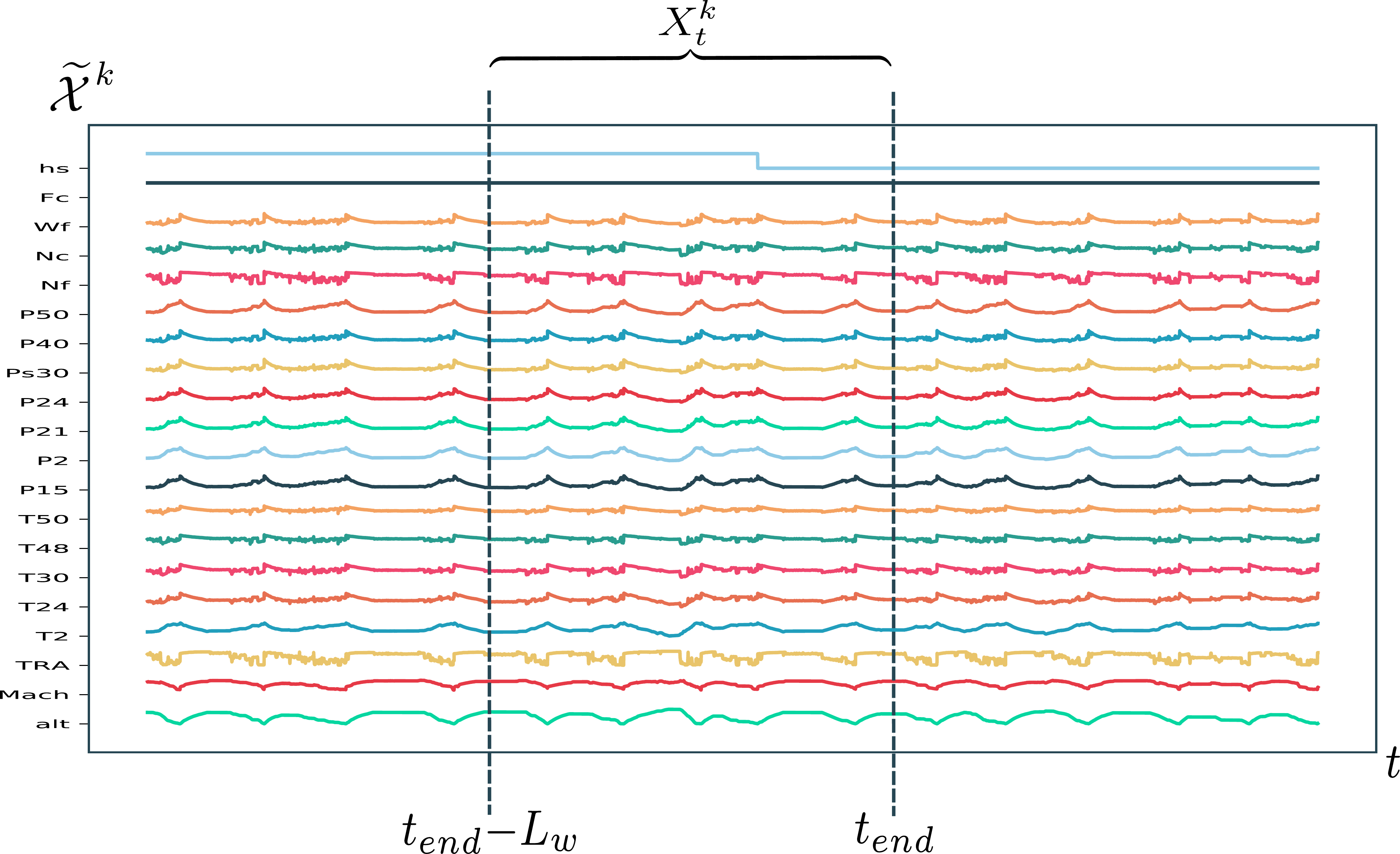}
\caption{Sliding window.}
\label{fig:window}
\end{figure}

The network inputs are generated by sliding a time window through the normalized data, with the window size denoted as $L_w$ and determined during model selection. The inputs are defined as 
$$X^k_t = [\widetilde{\mathcal{X}}^k_{t_{end} - L_w}, ..., \widetilde{\mathcal{X}}^k_{t_{end}}]$$
\noindent where $t_{end}$ is the end time of the window (see Figure \ref{fig:window}). The corresponding ground truth RUL label for each input is denoted as $Y_t$. This method generates $T^k - L_w$ samples for each unit, where $T^k$ represents the total run time in seconds of the unit.

The ground-RUL label has been defined as a linear function of cycles from the RUL of each unit $Y^k_t = TUL^k -  C^k_t$, where $TUL^k$ is the total useful life of the unit $k$ in cycles and $C^k_t$ is the number of past cycles from the beginning of the experiment at time $t$.

\begin{figure}[t]
\centering
\includegraphics[scale=.13]{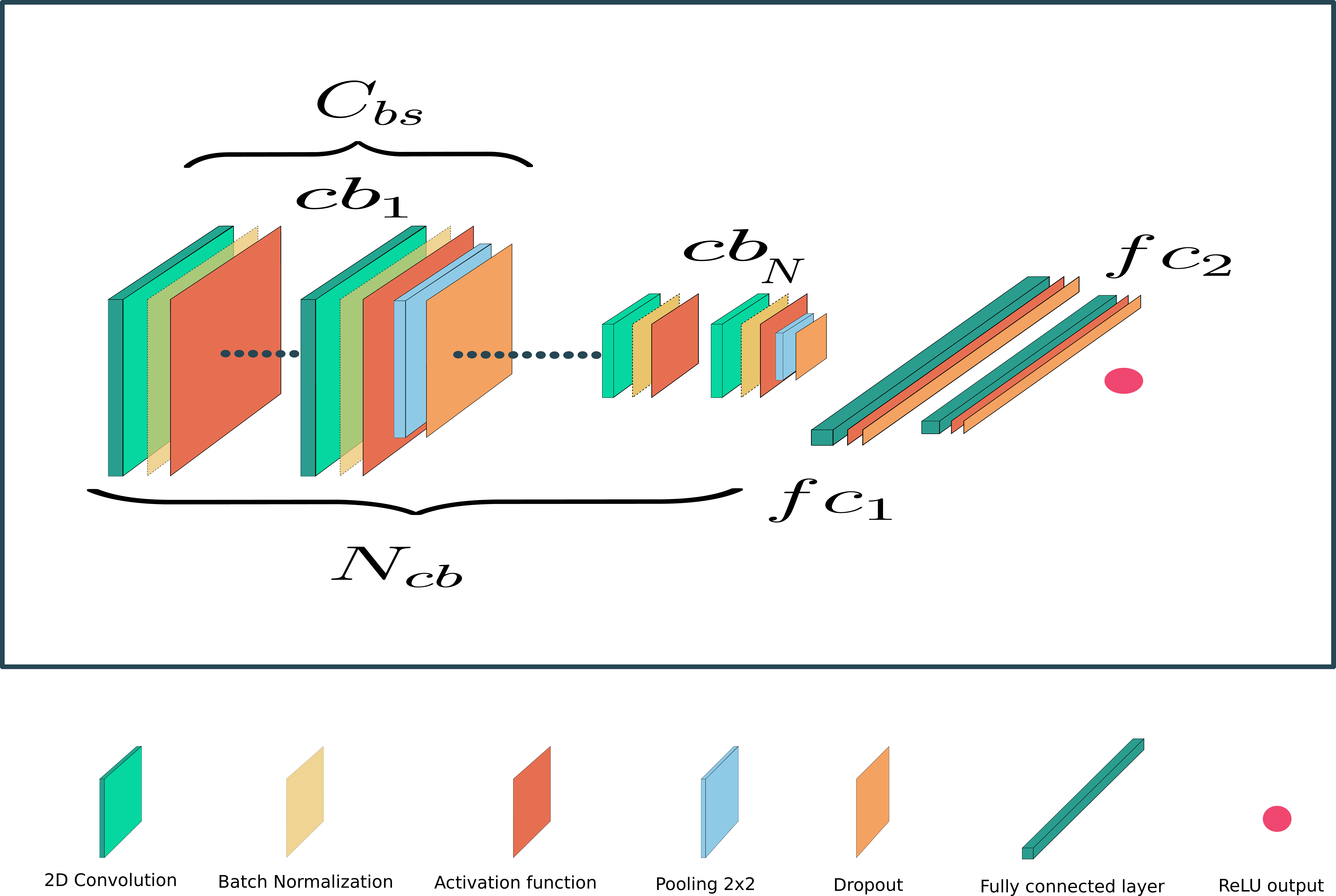}
\caption{DCNN network architecture used in experiments.}
\label{fig:architectures}
\end{figure}

The black-box model is a Deep Convolutional Neural Network (DCNN). The classical DCNN architecture, which is shown in Figure \ref{fig:architectures}, can be divided into two parts. The architecture is composed of two main parts. The first part consists of a stacking of $N_b$ blocks that include convolutional and pooling layers. The main objective of this part is to extract relevant features for the task at hand. The second part is made up of fully connected layers, which are responsible for performing the regression of the RUL in this case. The detailed parameters of the network can be found in Table \ref{table:param_ranges}. The network is composed of four convolutional blocks, each of which consists of four stacked convolutional layers with hyperbolic tangent ($\tanh$) activation functions. Following the convolutional blocks, two fully connected layers with leaky ReLU activation are stacked. The output layer uses the rectified linear unit (ReLU) activation function, since negative values in the output are not desired.

\begin{algorithm}
\caption{Algorithm to compute each proxy on the test set}\label{alg:cap}
\begin{algorithmic}
\State $X$ is a set of $N$ samples
\State $P$ is a proxy
\State $N \gets \left | X \right |$
\State $S \gets 0$
\For {$x_i \in X$}
\State $s_i \gets P(x_i)$
\State $S \gets S + s_i$
\EndFor
\State \Return $\frac{S}{N}$
\end{algorithmic}
\label{algo:scoring}
\end{algorithm}

\begin{table}[]
\centering
\caption{Parameters of the black-box model developed}
\begin{tabular}{lcrr}
\toprule
\textbf{Patameter}            & \textbf{Value}      \\ 
\midrule
$L_w$                      & 161         \\      
\midrule
$B_s$                        & 116            \\   \hline
$C_{bs}$                       & 4                        \\ \hline
$N_{cb}$                       & 4              \\ \hline
$fc_1$                       & 256     \\ \hline
$fc_2$                       & 100     \\ \hline
$\sigma_{conv}$            & tanh  \\ \hline
$d_{rate}$                 & 2   \\ \hline
$\sigma_{fc}$              &  Leaky ReLU      \\ \hline
$\sigma_{output}$              &  ReLU  \\ 
\toprule
\#Net params     &  1,514,016    \\ \hline
RMSE               &    10.46    \\ \hline
MAE               &    7.689           \\ \hline
NASA score          &  2.13             \\ \hline
CV \: $\mathcal{S}$ score  & 6.30  \\ \hline
std($\mathcal{S}$)         &   0.37 \\ \hline
\bottomrule
\end{tabular}
\label{table:param_ranges}
\end{table}

\subsection{Experiments}

The experiments were conducted using a dataset consisting of 256 data samples. These samples were not used during the model training phase. Each data sample contains 20 time series, and each time series consists of 160 time units. The 8 proxies, that were defined in Section \ref{sect:proxies}, were computed for each sample, and the final score for each proxy was determined by taking the mean of the 256 samples as it is described in the algorithm \ref{algo:scoring}. This process was repeated for each XAI method studied.

\begin{figure}[t]
\centering
\includegraphics[scale=.2]{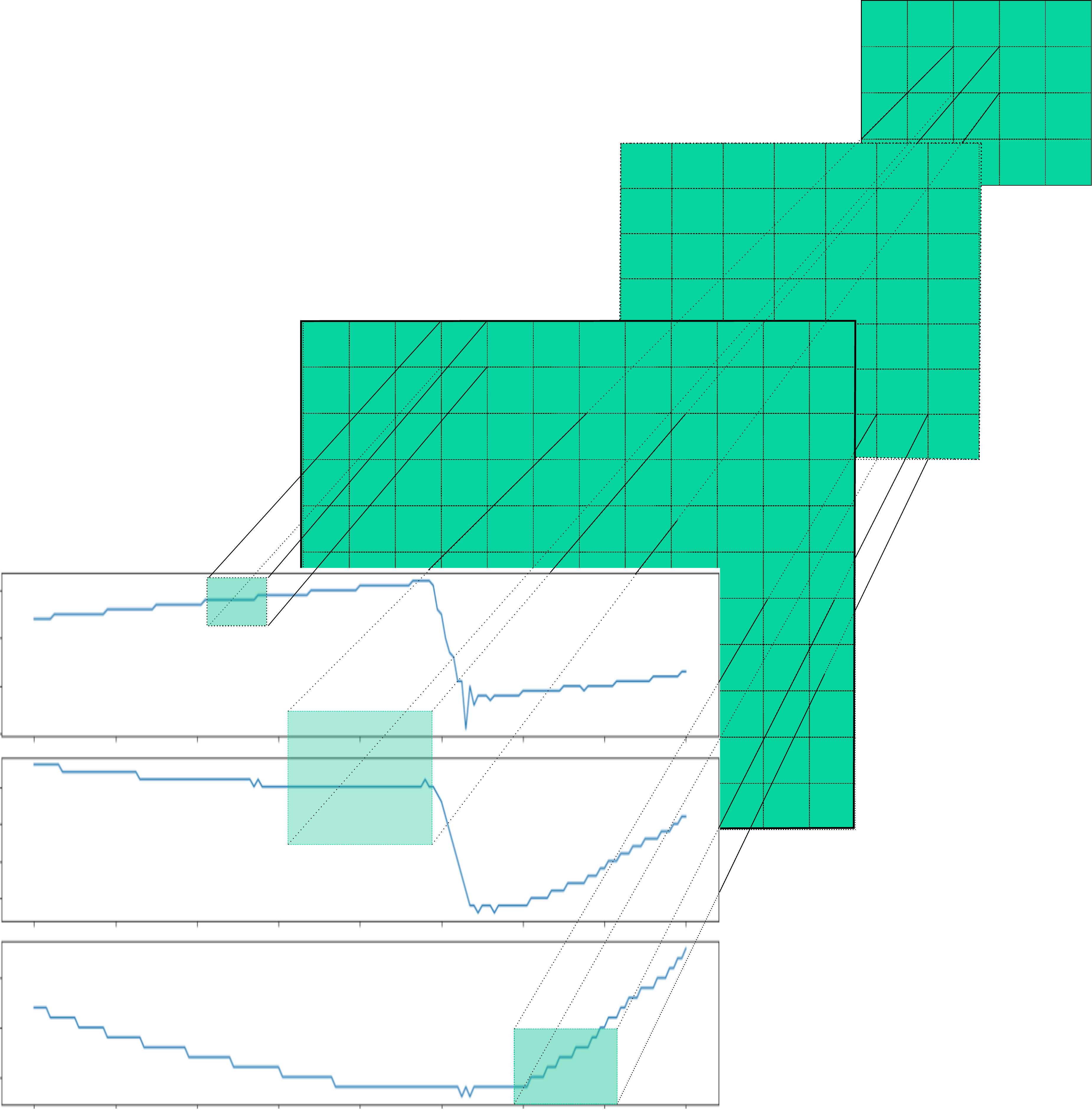}
\caption{Grad-CAM feature map distribution over the input depending on its depth within the network}
\label{fig:gradcam-layers}
\end{figure}

For both LIME and SHAP, five perturbation methods were tested: zero, one, mean, uniform noise, and normal noise. In the case of selectivity, due to performance issues, groups of 10 features, ordered by importance, were considered for computing the AUC. For the remaining XAI methods, the replace-by-zero perturbation method has been used since, in the experiments performed, it provided the best results. Finally, in Grad-CAM, different values between 0 and 1 for $\beta$ and $\sigma$ (factors adjusting the contribution of the time and feature components in the computation of feature importance) were tested using a grid search methodology with a step of 0.1. Furthermore, heat maps were extracted for each convolutional layer to study the best layer for explaining the predictions of the DCNN model on this dataset. It is important to note that the gradients are distributed differently depending on its depth within the network (Figure \ref{fig:gradcam-layers}), which means that the last convolutional layer, commonly exploited in the literature, may not be the best to solve all problems. By using the proxies, it is possible to assess which is the best layer from the perspective of explainability.

\subsection{Results}

The results for each method are presented in Table \ref{table:rmethods}. This table shows that Grad-CAM achieved the highest value for 5 out of the 8 proxies tested, and for selectivity and completeness, the value was close to the maximum. The best result for Grad-CAM was obtained using $\beta = 0.9$, $\sigma = 0.0$, and computing the gradient with respect to the second layer of the network. Note that Grad-CAM gets the worst result, compared to the rest of the proxies, according to the \textit{acumen} proxy, defined in this work. 

\begin{table}[H]
\centering
\caption{This table shows the result of the models for the different proxies. Perm: Permutation, I: Identity, Sep: Separability, Sel: Selectivity, Sta: Stability, Coh: Coherence, Comp: Completeness, Cong: Congruency, Acu: Acumen\label{identity-results}}
\begin{tabular}{lcrrrrrrrr}
\toprule
\textbf{Method}      & \textbf{Perm} & \textbf{I} & \textbf{Sep} & \textbf{Sta} & \textbf{Sel} & \textbf{Coh} & \textbf{Comp} & \textbf{Cong} & \textbf{Acu} \\
\midrule
\textbf{Saliency}    &                      & 1.000               & 0.999                 & 0.055              & 0.450                 & 0.174              & 0.972                 & 0.163               & 0.516           \\
\midrule
\textbf{LRP}         &                      & 1.000               & 1.000                   & -0.037             & 0.599                & 0.180               & 0.967                 & 0.165               & 0.495           \\
\midrule
\textbf{Lime}        & mean                 & 0.004             & 1.000                   & 0.130               & 0.573                & 0.173              & 0.962                 & 0.161               & 0.685           \\
\textbf{Lime}        & n. noise        & 0.008             & 1.000                   & 0.131              & 0.582                & 0.173              & 0.960                  & 0.162               & 0.677           \\
\textbf{Lime}        & u. noise       & 0.012             & 1.000                   & 0.109              & 0.560                & 0.166              & 0.960                  & 0.162               & 0.577           \\
\textbf{Lime}        & zero                 & 1.000               & 1.000                   & 0.554              & \textbf{0.835}                 & 0.16               & \textbf{1.017}                 & 0.146               & 0.753           \\
\textbf{Lime}        & one                  & 1.000               & 1.000                   & 0.349              & 0.728                & 0.184              & 0.969                 & 0.166               & 0.069           \\
\midrule
\textbf{Grad-CAM}     &                      & \textbf{1.000}               & \textbf{1.000}                   & \textbf{0.653}              & 0.741                & \textbf{0.196}              & 0.947                 & \textbf{0.171}               & 0.435           \\
\midrule
\textbf{SHAP} & mean                 & 0.000               & 1.000                   & 0.033              & 0.582                & 0.120               & 0.973                 & 0.152               & 0.505           \\
\textbf{SHAP} & n. noise        & 0.000               & 1.000                   & 0.037              & 0.581                & 0.116              & 0.968                 & 0.150                & 0.501           \\
\textbf{SHAP} & u. noise       & 0.000               & 1.000                   & 0.027              & 0.581                & 0.125              & 0.961                 & 0.162               & 0.503           \\
\textbf{SHAP} & zero                 & 1.000               & 1.000                   & 0.226              & 0.800                & 0.152              & 1.001                  & 0.149               & \textbf{0.761}           \\
\textbf{SHAP} & one                  & 1.000               & 1.000                   & 0.200                & 0.692                & 0.173              & 0.969                 & 0.169               & 0.349   \\ 
\bottomrule
\end{tabular}
\label{table:rmethods}
\end{table}

\begin{figure}[t]
\centering
\includegraphics[scale=.4]{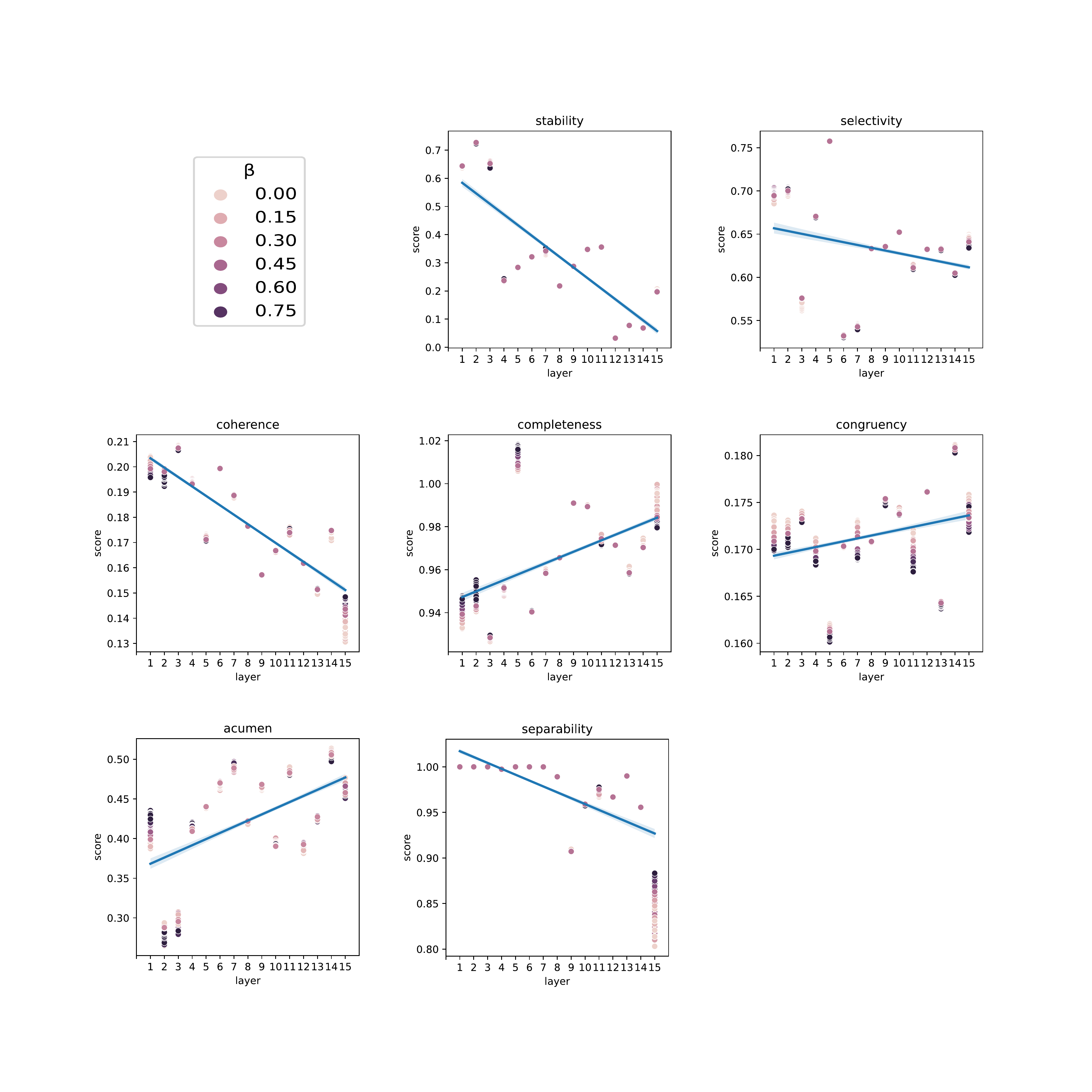}
\caption{Grad-CAM behavior for each proxy in each layer}
\label{fig:gardcam_layers}
\end{figure}

Since Grad-CAM is clearly the method providing the best results, further analysis have been carried out to understand the behavior of the method, under different settings, for each of the proxies studied. The identity proxy was omitted, as it is always 1 in Grad-CAM. Figure \ref{fig:gardcam_layers} compares the scores obtained by Grad-CAM when applied to each convolutional layer of the DCNN. The selectivity, stability, separability, and coherence proxies show an inverse correlation with respect to the depth of the layer. Conversely, the proxies for completeness, congruency and acumen present direct correlation with the layer depth. 

\begin{figure}[t]
\centering
\includegraphics[scale=.4]{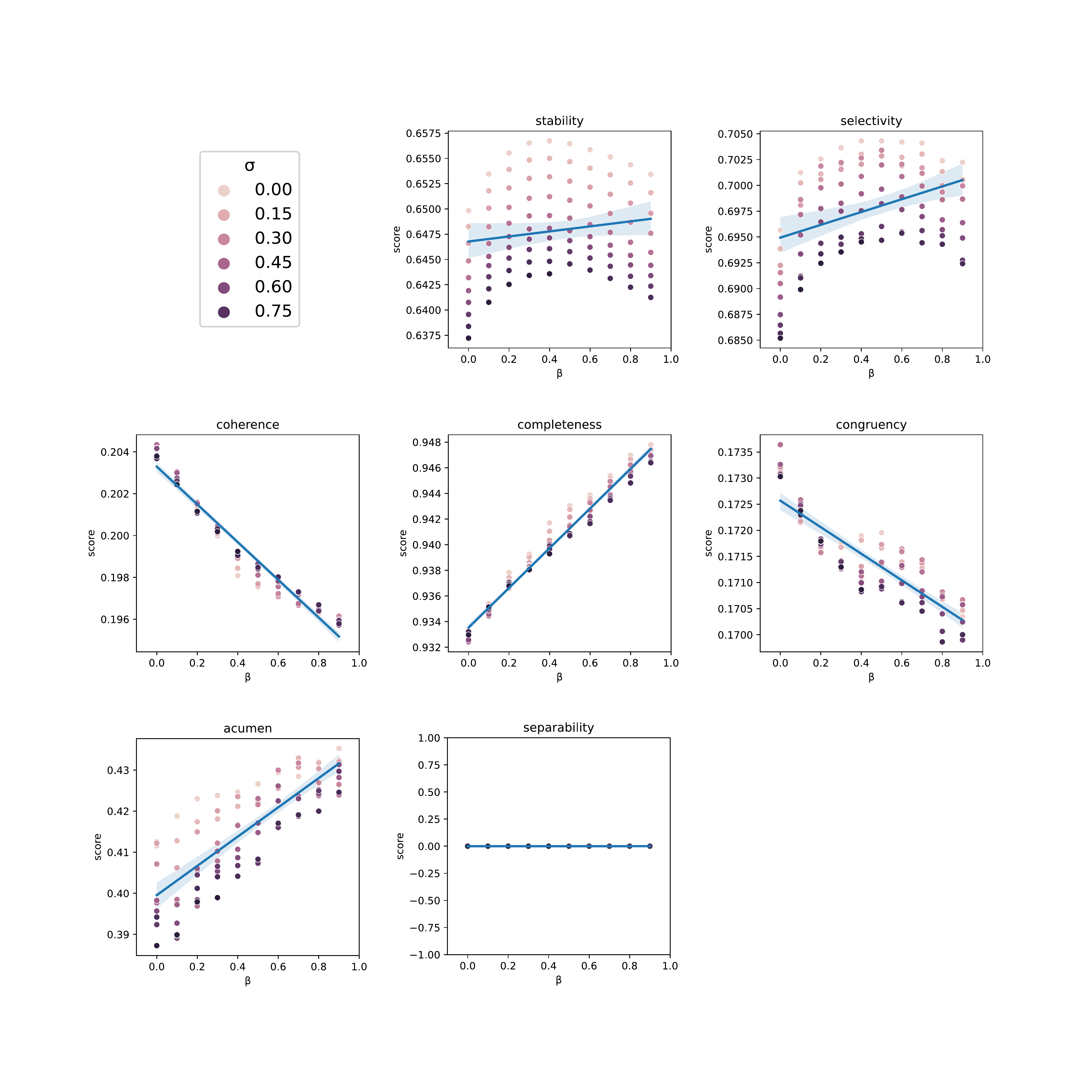}
\caption{Grad-CAM behavior for each proxy respect $\beta$}
\label{fig:gardcam_beta}
\end{figure}

The inverse correlation may be due to the existence of large groups of features having a higher likelihood of including features that impact the proxy negatively. For example, in the case of selectivity, a group that is considered important as a whole could contain samples with low importance. Therefore, for these proxies it is better to consider features independently, instead of as a group.
Figure \ref{fig:gardcam_beta} shows the influence of the time contribution, which is controlled by the factor $\beta$. The factor being discussed shows a direct correlation with all proxies excepting congruency and coherence, which present an inverse correlation. The reason for this effect on coherence may be due to the introduction of the time contribution, which spreads the importance across a larger number of features. As a consequence of this, the limit in considering a maximum of 100 features impacts negatively on these two proxies. Lastly, the feature contribution, weighted by the factor $\sigma$, was found to have no contribution in improving any of the proxy scores.

\section{Discussion}


This work is focused on the under-researched area of XAI methods for time series and regression problems. The first aim of this work was to review existing papers on XAI addressing such topics, with an emphasis in the use of quantitative metrics to compare XAI methods. Then, a comparison among the most promising XAI methods was carried out on a highly complex model, as is the DCNN, applied to a time series regression problem within the context of PHM. With this aim, a number of experiments were performed, quantifying the quality of explanations by computing eight different proxies. Results showed that Grad-CAM was the most effective XAI method among the ones tested, achieving the highest values for five out of eight proxies and being close to the maximum in two others. The results also showed that the impact of the layers and the time component contribution $\beta$ on Grad-CAM varied for different proxies, showing some of them a direct correlation and others an inverse correlation. The findings of this study highlight the need for further research in this area and the importance of developing better XAI methods for time series and regression problems, particularly in the PHM field.


In addition to comparing various XAI methods through quantitative proxies, this work also makes a significant contribution by introducing a new quantitative proxy called acumen, which measures a desirable property of any XAI method and highlights the breach of this property by Grad-CAM. Furthermore, an extension of Grad-CAM that takes into account time dependencies is proposed (where such contribution can be modulated), and results showed that this extension improves the performance of most of the studied XAI methods. These findings demonstrate the importance of considering time dependencies when evaluating the performance of XAI methods in time series and provide valuable insights for future research in this area.

The experiments were carried out on a deep neural network trained to predict the remaining useful life of a turbine engine, which belongs to a type of regression problem that has received little attention in XAI. All the code to reproduce the experiments, along with the data and the model, is provided to allow the research community to further explore these findings. Overall, this work makes a valuable contribution to the field of XAI by addressing important gaps in the literature and presenting novel approaches for time series and regression problems.


\begin{thebibliography}{999}


\bibitem[Pomerleau(1993)]{ref-pomerleau}
Pomerleau, Dean A. {\bf Neural networks for intelligent vehicles}. {\em Proceedings of IEEE Conf. on Intelligent Vehicles}, 19--24

\bibitem[Goodman(2017)]{ref-goodman2017european}
Goodman, Bryce and Flaxman, Seth. {\bf European Union regulations on algorithmic decision-making and a “right to explanation}. {\em AI magazine} {\bf 2017}, {\em 38}, 50--57

\bibitem[Arrieta(2020)]{ref-arieta}
Arrieta, Alejandro Barredo and D{\'\i}az-Rodr{\'\i}guez, Natalia and Del Ser, Javier and Bennetot, Adrien and Tabik, Siham and Barbado, Alberto and Garc{\'\i}a, Salvador and Gil-L{\'o}pez, Sergio and Molina, Daniel and Benjamins, Richard and others. {\bf Explainable Artificial Intelligence (XAI): Concepts, taxonomies, opportunities and challenges toward responsible AI}. {\em Information fusion} {\bf 2020}, {\em 58}, 82--115

\bibitem[Ribeiro(2016)]{ref-ribeiro}
Ribeiro, Marco Tulio and Singh, Sameer and Guestrin, Carlos. {\bf " Why should i trust you?" Explaining the predictions of any classifier}. {\em Proceedings of the 22nd ACM SIGKDD international conference on knowledge discovery and data mining} {\bf 2016}, 1135--1144

\bibitem[Lundberg(2017)]{ref-lundberg}
Lundberg, Scott M and Lee, Su-In. {\bf A unified approach to interpreting model predictions}. {\em Advances in neural information processing systems} {\bf 2017}, {\em 30}


\bibitem[Selvaraju(2017)]{ref-selvaraju}
Selvaraju, Ramprasaath R and Cogswell, Michael and Das, Abhishek and Vedantam, Ramakrishna and Parikh, Devi and Batra, Dhruv. {Grad-cam: Visual explanations from deep networks via gradient-based localization}. {\em Proceedings of the IEEE international conference on computer vision} {\bf 2017}, 618--626

\bibitem[Simonyan(2013)]{ref-simonyan}
Simonyan, Karen and Vedaldi, Andrea and Zisserman, Andrew. {\bf Deep inside convolutional networks: Visualising image classification models and saliency maps}. {\em arXiv preprint arXiv:1312.6034} {\bf 2013}


\bibitem[Bach(2015)]{ref-bach}
Bach, Sebastian and Binder, Alexander and Montavon, Gr{\'e}goire and Klauschen, Frederick and M{\"u}ller, Klaus-Robert and Samek, Wojciech. {\bf On pixel-wise explanations for non-linear classifier decisions by layer-wise relevance propagation}. {\em PloS one} {\bf 2015}, {\em 10(7)}, e0130140

\bibitem[Carvalho(2019)]{ref-carvalho}
{\em Carvalho, Diogo V and Pereira, Eduardo M and Cardoso, Jaime S} {\bf Machine learning interpretability: A survey on methods and metrics}, {\em Electronics}, {\bf 2019}, {\em 8(8)}, 832

\bibitem[Honegger(2018)]{ref-honegger}
{\em Honegger, M.} {\bf Shedding Light on Black Box Machine Learning Algorithms: Development of an Axiomatic Framework to Assess the Quality of Methods that Explain Individual Predictions}. {\em arXiv 2018, arXiv:1808.05054}

\bibitem[Doshi(2017)]{ref-doshi}
{\em Doshi-Velez, F.; Kim, B.} {\bf Towards a rigorous science of interpretable machine learning}. {\em arXiv 2017, arXiv:1702.08608}



\bibitem[Silva(2018)]{ref-silva}
{\em Silva, Wilson and Fernandes, Kelwin and Cardoso, Maria J and Cardoso, Jaime S} {\bf Towards complementary explanations using deep neural networks}. Understanding and Interpreting Machine Learning in Medical Image Computing Applications, {\bf 2018}; pp. 133--140.

\bibitem[Gilpin(2018)]{ref-gilpin}
{\em Gilpin, L.H.; Bau, D.; Yuan, B.Z.; Bajwa, A.; Specter, M.; Kagal, L.} {\bf Explaining explanations: An overview of interpretability of machine learning.}, {\em 2018 IEEE 5th International Conference on data science and advanced analytics (DSAA)}, {\bf 2018}; pp 80--89


\bibitem[Letzgus(2021)]{ref-letzgus}
{\em Letzgus, Simon and Wagner, Patrick and Lederer, Jonas and Samek, Wojciech and M{\"u}ller, Klaus-Robert and Montavon, Gr{\'e}goire} {\bf Toward Explainable AI for Regression Models}, {\em IEEE Signal Processing Magazine}, {\bf 2022}, {\em 39(4)};  pp 40--58


\bibitem[Schlegel(2019)]{ref-schlegel}
{\em Schlegel, Udo and Arnout, Hiba and El-Assady, Mennatallah and Oelke, Daniela and Keim, Daniel A} {\bf Towards a rigorous evaluation of xai methods on time series}, {\em 2019 IEEE/CVF International Conference on Computer Vision Workshop (ICCVW)}, {\bf 2019}; 4197--4201

\bibitem[Samek(2017)]{ref-samek}
{\em Samek, Wojciech and Wiegand, Thomas and M{\"u}ller, Klaus-Robert} {\bf Explainable artificial intelligence: Understanding, visualizing and interpreting deep learning models}, {\em arXiv preprint arXiv:1708.08296}, , {\bf 2017}

\bibitem[Samek(2016)]{ref-samek2}
{\em Samek W, Binder A, Montavon G, Lapuschkin S, Müller KR}. {\bf Evaluating the visualization of what a deep neural network has learned}. {\em IEEE transactions on neural networks and learning systems}. {\bf 2016}, 28(11):2660-73.

\bibitem[Siddiqui(2019)]{ref-siddiqui}
{\em Siddiqui, Shoaib Ahmed and Mercier, Dominique and Munir, Mohsin and Dengel, Andreas and Ahmed, Sheraz} {\bf Tsviz: Demystification of deep learning models for time-series analysis}, {\em IEEE Access}, {\bf 2019}, {\em 7}, pp 67027--67040

\bibitem[Vollert(2021)]{ref-vollert}
{\em Vollert, Simon and Atzmueller, Martin and Theissler, Andreas} {\bf Interpretable Machine Learning: A brief survey from the predictive maintenance perspective}, {\em 2021 26th IEEE international conference on emerging technologies and factory automation (ETFA)}, {\bf 2021}, pp 01--08

\bibitem[Hong(2020)]{ref-hong}
{\em C. W. Hong, C. Lee, K. Lee, M.-S. Ko and K. Hur} {\bf Explainable Artificial Intelligence for the Remaining Useful Life Prognosis of the Turbofan Engines}, {\em 2020 3rd IEEE International Conference on Knowledge Innovation and Invention (ICKII)}, {\bf 2021},  pp. 144-147

\bibitem[Szelazek(2020)]{ref-szelazek}
{\em M. Szelazek, S. Bobek, A. Gonzalez-Pardo and G. J. Nalepa}, {\bf Towards the Modeling of the Hot Rolling Industrial Process. Preliminary Results}, {\em Intelligent Data Engineering and Automated Learning - IDEAL 2020, Cham: Springer}, {\bf 2020}; vol. 12489, pp. 385-396

\bibitem[Serradilla(2021)]{ref-serradilla}
{\em O. Serradilla, E. Zugasti, C. Cernuda, A. Aranburu, J. R. de Okariz and U. Zurutuza}, {\bf Interpreting Remaining Useful Life estimations combining Explainable Artificial Intelligence and domain knowledge in industrial machinery}, {\em 2020 IEEE International Conference on Fuzzy Systems (FUZZ-IEEE)}, {\bf 2020}, pp. 1-8

\bibitem[Ferrano(2022)]{ref-ferrano}
{\em Ferraro A, Galli A, Moscato V, Sperlì G.}, {\bf Evaluating eXplainable artificial intelligence tools for hard disk drive predictive maintenance}. {\em Artificial Intelligence Review}, {\bf 2022}, pp 1-36

\bibitem[Shapley(1953)]{ref-shapley}
{\em Shapley LS}. {\bf A value for n-person games}, {\bf 1953}

\bibitem[Zhou(2016)]{ref-zhou}
{\em B. Zhou, A. Khosla, L. A., A. Oliva, and A. Torralba}. {\bf Learning Deep Features for Discriminative Localization}, {\em CVPR}, {\bf 2016}

\bibitem[Truong(2020)]{ref-truong}
{\em C. Truong, L. Oudre, N. Vayatis}. {\bf Selective review of offline change point detection methods}, {\em Signal Processing, 167:107299}, {\bf 2020}. 

\bibitem[Arias(2021)]{ref-arias}
{\em Arias Chao, M., Kulkarni, C., Goebel, K., Fink, O.}, {\bf Aircraft engine run-to-failure dataset under real flight conditions for prognostics and diagnostics}. {\em Data}, {\bf 2021}, 6(1), p 5

\bibitem[Saxena(2008)]{ref-saxena}
{\em Saxena A, Goebel K, Simon D, Eklund N.} {\bf Damage propagation modeling for aircraft engine run-to-failure simulation}, {\em 2008 international conference on prognostics and health management. IEEE}, {\bf 2008 O}, pp. 1-9

\bibitem[Rokade(2008)]{ref-rokade}
{\em Rokade P, BKSP KR}. {\bf Building Quantifiable System for Xai Models}, {\em SSRN 4038039}

\bibitem[Solis(2021]{ref-solis}
{\em Solís-Martín D, Galán-Páez J, Borrego-Díaz J. A}  {\bf Stacked Deep Convolutional Neural Network to Predict the Remaining Useful Life of a Turbofan Engine}. {\em Annual Conference of the PHM Society}, {\bf 2021}, Vol. 13, No. 1.

\end{thebibliography}
\end{document}